%% file: spellchecker.tex
\pdfoutput=1 

\documentclass[11pt]{article}

\usepackage[]{EACL2023} 

\usepackage{times}
\usepackage{latexsym}

\usepackage[T1]{fontenc} 

\usepackage[utf8]{inputenc} 

\usepackage{hyperref}
\hypersetup{pdftitle={Wolof Spellchecking}}

\usepackage{tipa}

\usepackage{microtype} 

\usepackage{inconsolata} 

\usepackage{graphicx} 
\graphicspath{ {./figures/} }

\usepackage{tikz} 
\usetikzlibrary{shapes.geometric, arrows.meta, positioning}
\tikzstyle{box} = [rectangle, draw, text centered, rounded corners]
\tikzstyle{line} = [draw, -latex]

\usepackage{sepfootnotes} 

\usepackage{amsmath} 

\usepackage{multirow}

\title{Automatic Spell Checker and Correction for Under-represented Spoken Languages: Case Study on Wolof}

\author{Thierno Ibrahima Ciss{\'e} \\
  Universit{\'e} du Qu{\'e}bec {\`a} Montr{\'e}al \\
  \texttt{cisse.thierno{\_}ibrahima@courrier.uqam.ca} \\\And
  Fatiha Sadat \\
  Universit{\'e} du Qu{\'e}bec {\`a} Montr{\'e}al \\
  \texttt{sadat.fatiha@uqam.ca} \\}

\begin{document}

\maketitle

\input{footnotes}
\input{sections/abstract}
\input{sections/introduction}
\input{sections/sota}
\input{sections/methodology}
\input{sections/evaluation}
\input{sections/limitations}
\input{sections/conclusion.tex}

\input{sections/acknowledgements}

\bibliographystyle{acl_natbib}
\bibliography{asc}

\end{document}

%% file: footnotes.tex
\sepfootnotecontent{wolnews}    {https://www.wolof-online.com/}
\sepfootnotecontent{woltwi}        {https://twitter.com/SaabalN}
\sepfootnotecontent{wolrel}     {http://biblewolof.com/}
\sepfootnotecontent{tools}      {https://github.com/TiDev00/Wolof\_SpellChecker}

%% file: sections/abstract.tex
\begin{abstract}
This paper presents a spell checker and correction tool specifically designed for Wolof, an under-represented spoken language in Africa. The proposed spell checker leverages a combination of a trie data structure, dynamic programming, and the weighted Levenshtein distance to generate suggestions for misspelled words. We created  novel linguistic resources for Wolof, such as a lexicon and a corpus of misspelled words, using a semi-automatic 
approach that combines manual and automatic annotation methods. Despite the limited data available for 
the Wolof language, the spell checker's performance showed a predictive accuracy of 98.31\% and a 
suggestion accuracy of 93.33\%. 

Our primary focus remains the revitalization and preservation of Wolof as an Indigenous and spoken language in Africa, providing our efforts to develop novel linguistic resources. This work represents a valuable contribution to the growth of computational tools and resources for the Wolof language and provides a strong foundation for future studies in the automatic spell checking and correction field. 
\end{abstract}

%% file: sections/introduction.tex
\section{Introduction}

Linguistic diversity in Natural Language Processing (NLP) is essential to enable communication between different 
users and thus the development of linguistic tools that will serve the inclusion of diverse communities. 
Several research studies for low-resource languages have emerged; however spoken Indigenous and Endangered languages in Africa have been 
neglected, even though the cultural and linguistic richness they contain is inestimable.\\
The Wolof language, a popular language in west Africa spoken by almost 10 million individuals worldwide, is an immensely 
popular lingua franca in countries on the African continent, such as Senegal, Gambia and Mauritania.
It serves as the primary dialect of Senegal \citep{diouf2017}, hailing from the Senegambian branch of Niger-Congo's expansive language family. Furthermore, the language has been officially acknowledged in West Africa \citep{eberhard2019}. 
It is therefore not surprising that the intensive use of Wolof within the region has allowed it to be recognized as being of paramount importance.\\
Like several Indigenous and spoken languages of Africa, Wolof presents many challenges and issues, among which is the lack of linguistic resources and tools. Moreover, it is distinguished by its distinct tonal system which uses nasal vowels. The Wolof script is comprised of a total of 45 consonant phonemes, which are further subdivided into categories \citep{DicoWolFr2004}. Table \ref{tab:consonants} illustrates the various Wolof consonants and their respective classifications.
\begin{table}[ht]
\centering
\begin{tabular}{ccc}
\hline
\multicolumn{3}{c}{\textbf{Consonants}}                                                                              \\ \hline
\multicolumn{1}{c|}{\multirow{2}{*}{\textbf{Weak}}} & \multicolumn{2}{c}{\textbf{Strong}}                            \\ \cline{2-3} 
\multicolumn{1}{c|}{}                               & \multicolumn{1}{c|}{\textbf{Geminate}} & \textbf{Prenazalized} \\ \hline
\multicolumn{1}{c|}{\begin{tabular}[c]{@{}c@{}}p, t, c, \\ k, q, b, \\ d, j, g, \\ m, n, {\~n},\\ \textipa{\ng}, f, r,\\ s, x, w,\\ l, y\end{tabular}} &
  \multicolumn{1}{c|}{\begin{tabular}[c]{@{}c@{}}pp, tt, cc,\\ kk, bb, dd,\\  jj, gg, \textipa{\ng\ng}, \\ ww, ll, mm,\\ nn, yy, {\~n}{\~n}, qq\end{tabular}} &
  \begin{tabular}[c]{@{}c@{}}mp, nt, nc,\\ nk, nq, mb, \\ nd, nj, ng\end{tabular} \\ \hline
\end{tabular}
\caption{Wolof Consonants and Classifications}
\label{tab:consonants}
\end{table}

Furthermore, the Wolof writing system integrates a set of 17 vowel phonemes \citep{DicoWolFr2004} complementing the already existing 45 consonant phonemes. Table \ref{tab:vowels} provides an overview of the Wolof vowels and their respective classifications.

\begin{table}[ht]
\centering
\begin{tabular}{cc}
\hline
\multicolumn{2}{c}{\textbf{Vowels}} \\ \hline
\multicolumn{1}{c|}{\textbf{Short}} &
  \textbf{Long} \\ \hline
\multicolumn{1}{c|}{\begin{tabular}[c]{@{}c@{}}a, {\`a}, {\~a},\\ i, o, {\'o},\\ u, e, {\"e}, {\'e}\end{tabular}} &
  \begin{tabular}[c]{@{}c@{}}ii, uu, {\'e}{\'e},\\ {\'o}{\'o}, ee, oo, aa\end{tabular} \\ \hline
\end{tabular}
\caption{Wolof Vowels and Classifications}
\label{tab:vowels}
\end{table}

As writing becomes increasingly important due to our digital age, automatic spell checking plays a vital role in 
making sure written communications are both efficient and accurate. Despite the lack of standardization in their 
orthography, there has been a surge of interest to develop spell checkers for African Indigenous languages due to 
their growing importance in education, commerce, and diplomacy. Consequently, the development of 
spell checkers for these languages is slowly increasing.

Our main contribution in this paper, is the development of new resources for the Wolof language. Specifically, we have created a 
spell checker for the autocorrection of Wolof text, as well as a corpus of misspelled words that will enable 
researchers to evaluate the performance of future autocorrection systems. Additionally, we have developed a Wolof 
lexicon that can be leveraged for a range of tasks beyond autocorrection, such as neural machine translation, 
automatic speech recognition, etc. \\
The resources that have been developed over the course of this study are made publicly accessible on 
GitHub\sepfootnote{tools}, thereby enabling wider dissemination and facilitating the reproducibility of the research 
findings.

The remainder of our paper is structured as follows: in Section \ref{sec:sota}, we conduct a brief literature 
review and discuss some published studies. In Section \ref{sec:metho}, we describe our proposed methodology and the 
novel linguistic resources, we developped. In Section \ref{sec:eval}, we present results and evaluations of our 
study. In Section \ref{sec:limit}, we show the limitations of our system through an error analysis. Finally, 
section \ref{sec:conclu} concludes the paper and show some promising perspectives for the future.

%% file: sections/sota.tex
\section{Background} \label{sec:sota}

Spelling correction consists in suggesting valid words closer to a wrong one. In order to create an automatic 
spelling correction system, it is imperative to comprehend the root causes of spelling errors \citep{baba2012}.\\ 
Several studies about spelling errors have been done, with a notable contribution from \citep{mitton1996} who thoroughly analyzed different types of spelling mistakes for English and described methods to construct an automatic spelling correction system. 
\citep{kukich1992}, on the other hand, presented a survey on documented findings on spelling error patterns and categorized spelling errors into two groups:
\begin{itemize}
    \item Lexical errors: Result of mistakes applied to individual words, regardless of their context within a 
    sentence \citep{hacken2001}.
    \item Grammatical errors: Include both morphological and syntactical errors. Morphological errors involve 
    deficiencies in linguistic elements such as derivation, inflection, prepositions, articles, personal pronouns, auxiliary 
    verbs, and determiners. Syntactical errors result from issues in linguistic components, including passive voice, tense, 
    noun phrases, auxiliary verbs, subject-verb agreement, and determiners \citep{gayo2018}.
\end{itemize}
The causes of spelling errors are diverse and can stem from both cognitive and typographical sources \citep{peterson1980}. 
Cognitive errors arise when an individual lacks the proper understanding of the correct spelling of a word while 
typographical errors take place when incorrect keystrokes are made when typing. Literature in the field of spelling 
correction has typically approached these error types separately, with various techniques developed specifically to address 
each type \citep{kukich1992}.\\
Despite the significance of language processing, there has been a shortfall of focus on the creation of automatic spelling  correction tools for low resource languages especially. While some attempts have been made to apply standard automatic spelling correction techniques to a few African indigenous languages (\citealp{bantu2022}; \citealp{bambara2014}; \citealp{hausa2014}), no such efforts have been made for the Wolof language. As far as we are aware, the only research solely dedicated to the correction of Wolof is \citep{lo2016}, which provides an overview of the state of the art and outlines potential solutions for developing a tailored orthographic corrector for Wolof.
This research adopts commonly used approaches in the field and assesses the performance of our system using various known evaluation metrics. 

%% file: sections/methodology.tex
\section{Methodology} \label{sec:metho}

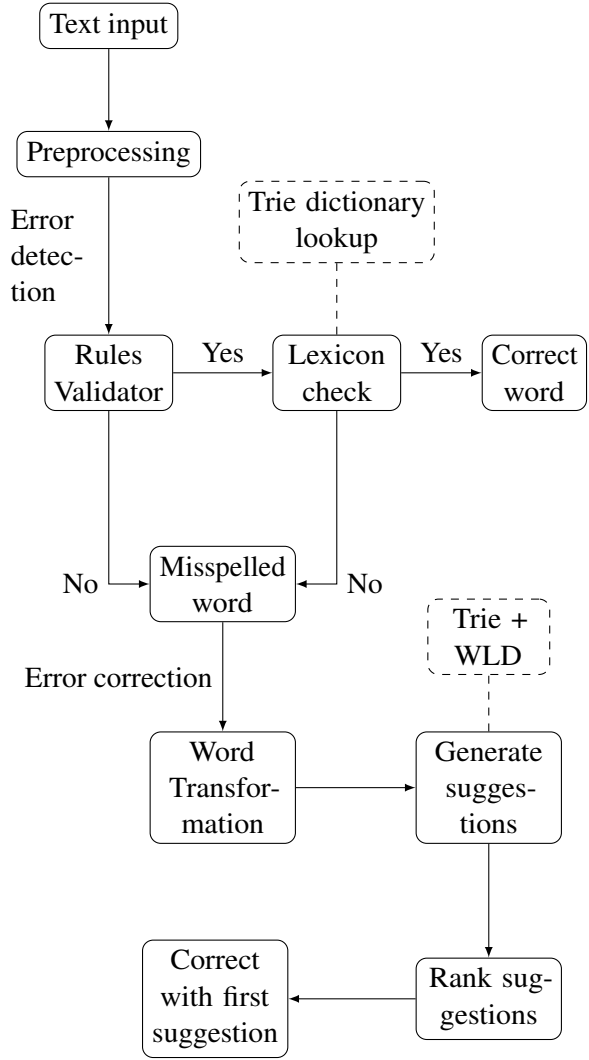
\begin{figure}[ht]
    \centering
    \begin{tikzpicture}[node distance=1.6cm]
        \node (input) [box] {Text input};
        \node (pp) [box, below of= input, yshift= -0.1cm] {Preprocessing};
        \node (rules) [box, below of= pp, text width= 3.7em, yshift = -1.3cm] {Rules Validator};
        \node (lookup) [box, right of= rules, text width= 3.7em, xshift= 1.4cm] {Lexicon check};
        \node (tree) [box, dashed, above of= lookup, text width= 6em, yshift= 0.4cm] {Trie dictionary lookup};
        \node (valword) [box, right of= lookup, text width= 2.9em, xshift= 1cm] {Correct word};
        \node (miss) [box, below of= rules, text width= 4.3em, yshift = -1.2cm, xshift= 1.5cm] {Misspelled word};
        \node (wt) [box, below of= miss, text width= 4.3em, yshift= -1.1cm] {Word Transformation};
        \node (wed) [box, right of= wt, text width= 4.3em, xshift= 1.9cm] {Generate suggestions};
        \node (trie) [box, dashed, above of= wed, text width= 3.5em, yshift= 0.4cm] {Trie + WLD};
        \node (rank) [box, below of= wed, text width = 4.3em, yshift = -1.2cm] {Rank suggestions};
        \node (disp) [box, left of= rank, text width = 4.3em, xshift= -2cm] {Correct with first suggestion};
    
        \path [line] (input) -- (pp);
        \path [line] (pp) -- node[left, text width = 3em] {Error detection} (rules);
        \path [line] (rules) -- node[above] {Yes} (lookup);
        \path [line, dashed, -] (lookup) -- (tree);
        \path [line] (lookup) -- node[above] {Yes} (valword);
        \path [line] (rules) |- node[left] {No} (miss);
        \path [line] (lookup) |- node[right] {No} (miss);
        \path [line] (miss) -- node[left] {Error correction} (wt);
        \path [line] (wt) -- (wed);
        \path [line, dashed, -] (wed) -- (trie);
        \path [line] (wed) -- (rank);
        \path [line] (rank) -- (disp);
    \end{tikzpicture}
    \caption{Flowchart of the spell checker}
    \label{fig:flow}
\end{figure}

The system outlined in this study aims to identify and correct non-word errors in Wolof
language. To achieve this objective, we designed and implemented the flowchart in Figure \ref{fig:flow}.

The flowchart in Figure \ref{fig:flow} provides a visual representation of the various components and processes 
involved in the proposed spell checker system. The system includes input and output mechanisms, algorithms for error detection and  correction, as well as data structures and models that aid in its functionality. The aim of the system is to  accurately identify and rectify non-word errors in the Wolof language. Through the presentation of the system's 
overall architecture, the reader will be able to comprehend the workings of the system and its design principles 
aimed at detecting and correcting invalid words in the Wolof language.

\subsection{Wolof Lexicon Generation}

Our approach to generating a reliable Wolof lexicon involves the combination of manual annotation 
and automatic extraction methods.\\
First, manual annotation was performed on a corpus of Wolof text \citep{nllb2022} to identify unique words and 
extract them into a list. This methodology provides a thorough examination of the Wolof language and ensures the 
precision of the lexicon by enabling manual control over the inclusion of words.\\
Second, an automatic extraction was performed using Optical Character Recognition (OCR) methods and implemented in 
the form of Python scripts, applied to several Wolof-French dictionaries (\citealp{DicoWolFr1990} and 
\citealp{DicoWolFr2001}). This methodology facilitates the expansion of the lexicon's coverage and enables the 
identification of additional words that may not have been captured through the manual annotation alone.\\
Finally, to ensure the lexicon's accuracy, the overall resulting data underwent proofreading. This step validated 
the correctness of the words and their spellings and allowed for any necessary revisions before the final lexicon was generated.\\
It is important to note that due to the limited availability of Wolof resources, the resulting lexicon only contains 1410 different words. Despite this constraint, the combination of manual annotation and automatic extraction methods allowed the generation of a reliable Wolof lexicon.

\subsection{Preprocessing}

Our spell checking system implements a preliminary stage, which entails the removal of inputs that contain 
numerical characters, punctuation marks, or borrowed words from foreign languages. The outcome of this step serves 
as the basis for the detection of non-word errors in the text.\\
The preprocessing phase consists of three primary operations: the elimination of punctuation marks, normalization 
of the input, and segmentation of the text into individual words.

\subsubsection{Punctuation Removal}

The goal of the punctuation removal in our preprocessing step is to eliminate any non-essential punctuation marks 
present in the input text. These marks can hinder the efficiency of the spell checking process and may cause 
confusion during the analysis of the text \citep{rahimi2022}. By removing these marks, we ensure that the 
subsequent stages of the spell checking system can process the text more effectively and efficiently. The algorithm 
employed in this stage scans the input text and removes all instances of punctuation marks, including commas, 
periods, exclamation marks, and question marks. The output of this step is a cleaned text that is free of the extraneous elements and ready for focused analysis in subsequent stages of the system.

\subsubsection{Normalization}

The normalization phase in our preprocessing step transforms the input text into a standardized form by converting all alphabetic characters to lowercase and removing any words outside of the Wolof language.\\
The conversion to lowercase is essential as many NLP techniques treat words in different cases as separate 
entities, and converting the text to lowercase eliminates this case sensitivity impact on the analysis 
\citep{kerner2020}.\\
By removing words from foreign languages, we aim to ensure that the text being analyzed is only in the target language and minimize the effect of words that may not hold semantic significance in the context of the Wolof language. This enhances the accuracy of the analysis and reduces the likelihood of introducing errors into the results.

\subsubsection{Word Tokenization}

Tokenization is a critical step in the automatic spell checking process, as it segments the input text into smaller units referred to as tokens.\\
There is a range of techniques used for tokenization that vary depending on the language and task at hand. These 
may include splitting on whitespace and punctuation, using regular expressions, or utilizing dictionaries and 
morphological rules \citep{dalry2006}.\\
The tokenization process results in units that can range from individual characters or words to phrases or even 
full sentences. However, word tokenization is the most commonly used form of segmentation in spell checking 
systems, as it separates the text into individual words and provides a solid foundation for the identification of 
spelling errors (\citealp{mosavi2014}; \citealp{hasan2021}; \citealp{abdul2022}).\\
Word tokenization not only enhances the accuracy and efficiency of the spell checking process but also allows for 
an analysis of the context in which each word appears. This enables the spell checking system to make more informed suggestions for appropriate spelling corrections, ultimately improving the accuracy of the results.\\ 
In the current investigation, we are implementing the process of tokenization with a focus on word-level segmentation.

\subsection{Error Detection}

The error detection stage in our spell checking system is designed to identify non-word errors in the text. This is 
achieved through a two-step process, consisting of validation against Wolof writing rules and comparison with a 
constructed lexicon. 

\subsubsection{Rules Validator}

The spelling of words in the Wolof language follows certain conventions, as described by the CVC and CVCV(C) forms 
for monosyllabic and disyllabic words, respectively \citep{merrill2021}. These conventions specify that the final 
consonant and vowel of a syllable cannot both be long, and strong consonants cannot appear after a long vowel or at 
the beginning of a word, except for prenasalized consonants.\\
Our error detection stage includes a validation step that rigorously checks each word in the input text against 
these writing conventions. If a word is found to be in compliance with these rules, it will move on to the next 
stage of validation. Conversely, if the word is determined to be non-compliant, it will be flagged as invalid and 
require correction.

\subsubsection{Lexicon Check}

In the lexicon verification phase, the spell checking system assesses each word in the input text against the Wolof 
lexicon to determine its validity. The lexicon, being a large repository of words, can pose challenges for quick 
and efficient searches. To address this, various techniques such as hash tables \citep{kukich1992}, binary search 
\citep{knuth1998}, tries data structure \citep{bentley1997}, and bloom filter \citep{bloom1970} have been developed 
to enable fast dictionary lookups.\\
In the present spell checker, the system uses the trie data structure, which organizes the lexicon into nodes that 
represent individual characters and the root node that represents the empty string. In this structure, searching 
for a word in the lexicon involves following the path through the trie that corresponds to the characters of the 
target word \citep{feng2012}. If the end of the path is  a terminal node, the word is considered to be in the 
lexicon and deemed valid. Conversely, if the path ends before reaching a terminal node, the word is considered 
incorrect and corrections are initiated.

\subsection{Error Correction}

The last stage of the spell correction procedure is to produce potential replacements for the incorrectly spelled 
word. Our correction techniques, described below, focus exclusively on the word and do not consider the context in 
which it appears. The correction process is comprised of three distinct phases: Translation of French Compound 
Sounds, Generation of Candidate Suggestions, and Ranking of those Suggestions.

\subsubsection{Translation of French compound sounds}

Prior to implementing the module responsible for translating French compound sounds into Wolof, we collected a small amount of Wolof data from various sources. This data was sourced from news websites\sepfootnote{wolnews}, social media platforms\sepfootnote{woltwi} and religious websites\sepfootnote{wolrel}. A thorough analysis of this data was carried out to determine the most common misspellings made by Wolof speakers when writing in the language. Our findings indicated that a significant number of these errors were due to the usage of the French alphabet instead of the Wolof alphabet. This often resulted in the presence of French compound sounds or letters that are not native to the Wolof language. Furthermore, it was observed that accents, which play a crucial role in ensuring proper pronunciation and meaning of words, were frequently neglected. To showcase these findings, Table 
\ref{tab:misspellings} presents some of the misspellings observed and their correct Wolof equivalent. 
\begin{table}[ht]
    \centering
    \begin{tabular}{c|c}
        \hline
        \textbf{Misspellings} & \textbf{Correct Wolof} \\
        \hline
        dadial{\'e} & dajale (to gather)\\
        guinaw & ginnaaw (behind)\\
        mousiba & musiba (danger)\\
        deuk & d{\"e}kk (village)\\
        thiossane & cosaan (tradition)\\
        gnopati & {\~n}oppati (to pinch)\\
        niaar & {\~naar} (two)\\
        sakhar & saxaar (train)\\
        tank & t{\`a}nk (foot)\\
        \hline
    \end{tabular}
    \caption{Misspellings and correct wolof words}
    \label{tab:misspellings}
\end{table}

Taking into consideration the common errors observed in the analysis of Wolof language data, our system is designed to assess each word for the presence of French compound sounds or letters that are extraneous to the Wolof alphabet. Should such sounds be detected, the module will translate them into their corresponding Wolof 
counterparts. Letters not belonging to the Wolof alphabet will be systematically eliminated. Upon completing these 
transformations, the output will be directed to the next phase of the correction process and candidate suggestions module.

\subsubsection{Generation of Candidate Suggestions}

In the current system, to generate potential alternatives for misspelled words, we have implemented a 
lexicographical distance comparison method. This process involves determining the minimum number of edit 
operations, such as insertion, deletion, transposition, and substitution, necessary to change one word into another 
\citep{vienney2004}. The more significant the disparities between two words, the greater the lexicographical 
distance between them. Out of various lexicographical distance metrics, the Levenshtein Distance \citep{lev1965} is 
the most commonly utilized. It quantifies the difference between two strings based on the three fundamental string 
operations: substitution, insertion, and deletion.\\
Let $Lev_{\alpha, \beta}$ be the Levenshtein distance between the subsequence formed with the $\alpha$ first characters of a word $W_1$ and the subsequence formed with the $\beta$ first characters of a word $W_2$. The Levenshtein distance between the two subsequences $W_1$ and $W_2$ (of length $|W_1|$ and $|W_2|$ respectively) can be recursively calculated using Formula \ref{form:lev} \citep{lev1965}.
\begin{equation}\label{form:lev}
    Lev_{\alpha,\beta} = \left\{\begin{array}{ll}
    \max(\alpha,\beta) \hspace{0.9cm} \textbf{if } \min(\alpha,\beta)=0\\
    \min\left\{ \begin{array}{l}
    lev_{\alpha-1, \beta} +1\\ 
    lev_{\alpha, \beta-1} +1\\ 
    lev_{\alpha-1, \beta-1} + 1_{(W_{1\alpha} \neq W_{2\beta})}
    \end{array}\right.
    \end{array}\right.
\end{equation}
\\The Levenshtein distance, computed using its recursive equation, can be computationally expensive 
\citep{benson1998}, especially for large distances as it has an exponential time complexity of $O(3^{min(|W_1|,|W_2|)})$. To address this issue, our approach combines two techniques: 
dynamic programming and the trie data structure.\\
Dynamic programming \citep{almu2001}, as a technique for solving problems by decomposing them into more manageable 
subproblems and storing the solutions, helps reduce the number of redundant calculations by providing a more 
efficient storage of intermediate results. When applied to the Levenshtein distance, it allows for the intermediate 
results of partial computations to be stored in a matrix, leading to a more efficient calculation of the final 
result.\\
By combining dynamic programming and the trie data structure, our approach effectively prunes the search space and 
avoids redundant calculations. This provides a powerful combination for computing the Levenshtein distance in a 
fast and efficient manner, even for large inputs.\\
In the standard Levenshtein distance, all edit operations are assigned a uniform cost of 1. However, considering 
the findings discussed earlier, a cost matrix was introduced to allow for the assignment of varying costs to 
different edit operations. This allows for a more nuanced representation of the importance of each operation. The 
cost for insertions and deletions remains at 1 for all characters. Substitution operations between source and 
target characters are assigned a cost of 1 if the character couple is listed in Table \ref{tab:costs}, otherwise, a cost of 2 is assigned.
\begin{table}[ht]
    \centering
    \begin{tabular}{c|c}
        \hline
        \textbf{Couple} & \textbf{Substitution cost} \\
        \hline
        ('a', '{\`a}') & 1\\
        ('a', '{\~a}') & 1\\
        ('o', '{\'o}') & 1\\
        ('e', '{\'e}') & 1\\
        ('e', '{\"e}') & 1\\
        ('{\'e}', '{\"e}') & 1\\
        ('x', 'q') & 1\\
        \hline
    \end{tabular}
    \caption{Substitution cost of specific couples}
    \label{tab:costs}
\end{table}

Our suggestion module generates potential candidate words for a given misspelled word through the computation of 
the edit distance between the misspelled word and each valid word in the Wolof lexicon. For each candidate word, 
the cost of transforming the misspelled word into the candidate word is provided.

\subsubsection{Ranking of candidate suggestions}

In the following phase of our methodology, the candidate words generated from the previous stage are subjected to 
evaluation. The ranking is performed based on the proximity of the candidate words to the misspelled word, with the 
candidate word having the smallest edit cost being assigned the highest rank. The candidate word with the lowest 
edit cost is determined to be the closest match and is therefore selected as the most likely substitution for the 
incorrect word.

%% file: sections/evaluation.tex
\section{Evaluations} \label{sec:eval}

In order to assess the performance of our spell checking system, we first constructed a corpus of misspelled words, then selected the appropriate evaluation metrics, and finally, we implemented the chosen metrics to assess the performance of the system.

\subsection{Generation of a Misspelled Word Corpus}

The creation of a Misspelled Word Corpus followed a similar method as the generation of the Wolof lexicon. We used 
a hybrid approach of manual and automatic annotation, followed by proofreading. The method involved the selection 
of commonly misspelled Wolof words discovered through social media, religious websites, and news websites. For each 
misspelling, we manually added its correction. This process resulted in the formation of a corpus consisting of 
3070 words, with 1075 valid words and 1995 invalid words. The edit distance between the misspelled words and their 
corrected forms is presented in Table \ref{tab:edit}.

\begin{table}[ht]
    \centering
    \begin{tabular}{c|c|c}
        \hline
        \textbf{Edit Distance} & \textbf{Count} & \textbf{Percentage}\\
        \hline
        1 & 400 & 20.05{\%} \\
        2 & 412 & 20.65{\%} \\
        3 & 445 & 22.31{\%} \\
        4 & 281 & 14.09{\%} \\
        5 & 204 & 10.23{\%} \\
        6 & 114 & 5.71{\%} \\
        7 & 67 & 3.36{\%} \\
        8 & 36 & 1.80{\%} \\
        9 & 23 & 1.15{\%} \\
        10 & 9 & 0.45{\%} \\
        11 & 2 & 0.1{\%} \\
        12 & 1 & 0.05{\%} \\
        13 & 1 & 0.05{\%} \\
        Total & 1995 & 100{\%} \\
        \hline
    \end{tabular}
    \caption{Edit distance of misspellings against their corrections}
    \label{tab:edit}
\end{table}

\subsection{Selection of the Evaluation Metrics}

There are various factors to consider in evaluating spelling checkers. Conventional metrics, including recall and 
precision, have been widely used for a considerable time to gauge the linguistic proficiency of such tools. 
nevertheless, from a usage-centered approach, these evaluation parameters have limitations due to the absence of certain variables intrinsic to the evaluation of spelling checkers in these metrics. 

To determine the reliability of our spell checker, we employed the metrics proposed in \citep{starlander2002}, \citep{voorhees2000}, \citep{Paggio1998}, \citep{Paggio1998a}, \citep{king1999} as well as the following measures:

\begin{itemize}
    \item True positive (TP): correct word which is recognized as correct by the spell checker.
    \item False positive (FP): incorrect word which is recognized as correct by the spell checker.
    \item False negative (FN): correct word which is recognized as incorrect by the spell checker.
    \item True negative (TN): incorrect word which is recognized as incorrect by the spell checker.
\end{itemize}

Despite their age, these metrics remain widely used in the current state of the art for evaluating spell checkers, 
particularly those designed for low-resource languages such as \citep{abdul2022} and \citep{bantu2022}.

The other used metrics in these evlautions, are described as follows:

\subsubsection{Lexical Recall or \boldmath{$R_c$}}

It is determined by calculating the ratio of correctly recognized valid words in the text by the spell checker, to 
the total number of accurate words in the same text, as shown in  Formula \ref{form:Rc} \citep{starlander2002}.

\begin{equation} \label{form:Rc}
    R_c=\frac{T_p}{T_p + F_n} 
\end{equation}

\subsubsection{Error Recall or \boldmath{$R_i$}}

It is expressed as the fraction of incorrect words in the text detected by the spell checker, compared to the overall number of incorrect words in the text, as shown in Formula \ref{form:Ri} \citep{starlander2002}.

\begin{equation} \label{form:Ri}
    R_i=\frac{T_n}{T_n + F_p}
\end{equation}

\subsubsection{Lexical Precision or \boldmath{$P_c$}}

It is calculated by dividing the total number of valid words accurately recognized by the spelling checker by the sum of valid words recognized by the spell checker and the quantity of invalid words that were not identified by the spell checker as incorrect, as shown in Formula \ref{form:Pc} \citep{starlander2002}.

\begin{equation} \label{form:Pc}
    P_c=\frac{T_p}{T_p + F_p}
\end{equation}

\subsubsection{Error Precision or \boldmath{$P_i$}}

It is determined by dividing the number of accurate flags made by the spell checker by the total number of flags 
issued by the system, as shown in Formula \ref{form:Pi} \citep{starlander2002}. 

\begin{equation} \label{form:Pi}
    P_i=\frac{T_n}{T_n + F_n}
\end{equation}

\subsubsection{Lexical F-measure or \boldmath{$Fm_c$}}

It enables the calculation of the harmonic mean between lexical recall and lexical precision, as shown in Formula \ref{form:Fmc} \citep{starlander2002}. 

\begin{equation} \label{form:Fmc}
    Fm_c=\frac{2}{\frac{1}{R_c} + \frac{1}{P_c}}
\end{equation}

\subsubsection{Error F-measure or \boldmath{$Fm_i$}}

The Error F-measure is calculated by computing the harmonic mean of lexical recall and lexical precision, as shown in Formula \ref{form:Fmi} \citep{starlander2002}.

\begin{equation} \label{form:Fmi}
    Fm_i=\frac{2}{\frac{1}{R_i} + \frac{1}{P_i}}
\end{equation}

\subsubsection{Predictive Accuracy or \boldmath{$PA$}}

It quantifies the probability of any word, whether correct or incorrect, being processed correctly by the spelling 
checker. It is calculated using Formula \ref{form:PA} \citep{starlander2002}. 

\begin{equation} \label{form:PA}
    PA=\frac{T_p + T_n}{T_p + T_n + F_p + F_n}
\end{equation}

\subsubsection{Suggestion Adequacy or \boldmath{$SA$}}

It measures the ability of our spell checker to suggest accurate spelling alternatives for a misspelled word. Let 
$S$ denote a proper recommendation for an incorrect word and $N$ represent the total number of misspelled words. 
The Suggestion Adequacy of our system is calculated using Formula \ref{form:SA} \citep{starlander2002}.

\begin{equation} \label{form:SA}
    SA=\frac{1}{N} \sum_{i=1}^{n} S_i
\end{equation}

\subsubsection{Mean Reciprocal Rank or \boldmath{$MRR$}}

As previously stated, our spell checker systematically selects the first word in the list of recommendations as the 
most likely substitution for the misspelled word. However, as selecting the initial option in the recommended list 
may not always be the appropriate choice, we will utilize the $MRR$ metric to assess the ranking methodology. Let 
$N$ be the total number of incorrect words and $Rank_{i,c}$ be the position of the correct suggestion in the list 
of suggestion for the $i^{th}$ misspelled word in $N$. The $MRR$ is computed using the Formula \ref{form:MRR} \citep{voorhees2000}.

\begin{equation} \label{form:MRR}
    MRR=\frac{1}{N} \sum_{i=1}^{N} \frac{1}{Rank_{i,c}}
\end{equation}

\subsection{Experiments}

\subsubsection{Results}

The results of our spell checker, as demonstrated in Table \ref{tab:results}, exhibit a remarkable level of 
proficiency in various aspects of spelling correction.

\begin{table}[ht]
    \centering
    \begin{tabular}{c|c|c}
    \hline
    \textbf{Metrics} & \textbf{Ratio} & \textbf{Percentage}\\
    \hline
    $R_c$ & 1023/1075 & 95.16\%\\
    $R_i$ & 1995/1995 & 100\%\\
    $P_c$ & 1023/1023 & 100\%\\
    $P_i$ & 1995/2047 & 97.46\%\\
    $Fm_c$ & 0.9752 & 97.52\%\\
    $Fm_i$ & 0.9871 & 98.71\%\\
    $PA$ & 3018/3070 & 98.31\%\\
    $SA$ & 1862/1995 & 93.33\%\\
    $MRR$ & 0.9604 & 96.04\%\\
    \hline
    \end{tabular}
    \caption{Performance measures of the spell checker}
    \label{tab:results}
\end{table}

The recall score of 95.16\% ($R_c$) and 100\% ($R_i$) depicts the comprehensive nature of the lexicon utilized by 
the spell checker, as well as its relatively unspoiled status.\\
The spell checker exhibits an exceptional level of precision, with a score of 100\% ($P_c$) and 97.46\% ($P_i$), 
indicating its reliability in accurately identifying spelling errors as well as valid words.\\
The F-measure scores of 97.52\% ($Fm_c$) and 98.71\% ($Fm_i$) demonstrate the spell checker's avoidance of 
simplistic strategies, thereby ensuring its efficiency.\\
The spell checker's suggestion accuracy ($SA$) score of 93.33\% attests to the suitability and veracity of the most 
probable alternative to the misspelled word presented to the end-user.\\
The mean reciprocal rank ($MRR$) score of 96.04\% highlights the quality of the ranking of suggestions presented by 
the spell checker.\\
Finally, the overall linguistic performance of the spell checker, as indicated by its predictive accuracy ($PA$) 
score of 98.31\%, is of a highly satisfactory nature.\\

\subsubsection{Errors analysis}

To fully understand and identify the linguistic limitations of our spell checker, we conducted an 
investigation into the edit distances of the misspelled words for which the system produced an incorrect 
suggestion. The outcome of this study is presented in Table \ref{tab:errors}.
\begin{table}[ht]
    \centering
    \begin{tabular}{c|c|c}
        \hline
        \textbf{Edit Distance} & \textbf{Count} & \textbf{Percentage}\\
        \hline
        1 & 4 & 3.01{\%} \\
        2 & 17 & 12.78{\%} \\
        3 & 32 & 24.06{\%} \\
        4 & 20 & 15.04{\%} \\
        5 & 22 & 16.54{\%} \\
        6 & 13 & 9.77{\%} \\
        7 & 10 & 7.52{\%} \\
        8 & 11 & 8.27{\%} \\
        9 & 3 & 2.26{\%} \\
        10 & 1 & 0.75{\%} \\
        Total & 133 & 100{\%} \\
        \hline
    \end{tabular}
    \caption{Edit distance of misspellings with wrong suggestions}
    \label{tab:errors}
\end{table}

After a thorough examination of the results displayed, we surprisingly noted that there was no significant linear 
correlation between the edit distance of a misspelled word and the probability of the spell checker generating 
incorrect suggestions. 
These findings are in line with those displayed in Table \ref{tab:edit}. The majority of 
words in our misspelled word corpus had an edit distance of 3, which increased the likelihood of the spell checker 
producing a wrong suggestion for misspelled words with an edit distance of 3. Additionally, as misspelled words 
with edit distances of 11, 12, and 13 were under-represented in our corpus, the spell checker's suggestions for 
these words were all accurate. This reinforces our conclusion that the higher the frequency of misspelled words 
with a specific edit distance, the greater the chances of the spell checker generating inaccurate suggestions for 
misspelled words with that same edit distance.

%% file: sections/limitations.tex
\section{Limitations} \label{sec:limit}

Despite the impressive performance and minimal processing time of our spell checker, it is important to acknowledge 
its limitations.\\
Firstly, the spell checker is restricted to the words included in the created Wolof lexicon and cannot recognize words outside of it.
Secondly, the weighted Levenshtein distance algorithm used may not always accurately reflect the likelihood of 
different types of errors, leading to potential inaccuracies in the suggestions.\\
Thirdly, the dynamic programming and trie data structures utilized may result in false positive suggestions due to a
lack of consideration for the semantic meaning of words.
Additionally, the computational cost of our approach can be substantial, particularly for larger lexicons or words 
with numerous possible corrections.
Finally, the lack of context awareness may result in missed errors or incorrect suggestions.

%% file: sections/conclusion.tex
\section{Conclusion} \label{sec:conclu}

This paper presented a novel spell checker for the Wolof language, that has demonstrated its potential, owing to 
its effective combination of the trie data structure, dynamic programming, and weighted Levenshtein distance 
algorithms. The hybrid approach of manual and automatic annotation enabled the construction of a comprehensive 
lexicon and a robust Misspelled Word Corpus, allowing for a robust evaluation of the spell checker's potential 
despite the limited data available for the language. Through these efforts, we hope to advance the state of NLP research for the Wolof language and contribute to preserving the linguistic heritage of African nations, ensuring that their distinct cultural expressions are protected for future generations.\\
The findings of this research provide compelling evidence of the viability of the spell checker for the Wolof 
language, opening avenues for further improvement and exploration. 

For future research, it would be of interest to 
study the effect of increasing the lexicon and Misspelled Word Corpus on the spell checker's performance. 
Furthermore, a comparison of the spell checker's performance with other spell-checking methods used in low-resource languages, such as the Indigenous African languages, could provide valuable insights into the strengths and weaknesses of the current approach. The integration of state-of-the-art techniques,taking into consideration the context, such as those based on machine learning and Deep Neural Networks, into the spell checker could also be explored to further enhance its capabilities.

%% file: sections/acknowledgements.tex
\section*{Acknowledgements} \label{acknowledgements}

We thank the anonymous reviewers for their helpful comments and feedback. \\
We also thank the participants of the Wolof community for giving their time, wisdom, and expertise.